\documentclass[10pt,twocolumn,letterpaper]{article}
\usepackage[T1]{fontenc}
\newif\ifanonymous
\anonymousfalse
\ifanonymous
  \usepackage[review]{cvpr}
\else
  \usepackage[pagenumbers]{cvpr}
\fi
\usepackage{graphicx,amsmath,amssymb,booktabs,multirow,array,xcolor}
\definecolor{cvprblue}{rgb}{0.21,0.49,0.74}
\usepackage[pagebackref,breaklinks,colorlinks,allcolors=cvprblue]{hyperref}
\def\confName{CVPR}
\def\confYear{2027}

\title{BDNet: Fast‑slow thinking is all you need}
\author{Chenzhi Zheng\textsuperscript{1}\quad
Zhuoran Zheng\textsuperscript{2}\thanks{Corresponding author.}\\
\textsuperscript{1}Jinan University\\
\textsuperscript{2}National University of Defense Technology}
\ifanonymous\else
\hypersetup{pdfauthor={Chenzhi Zheng,  Zhuoran Zheng}}
\fi
\begin{document}
\maketitle
\begin{abstract}
We present BDNet, a hierarchical image-classification backbone built around fast image features and slow statistical references. Its baseline deformation operator (BDO) transforms the deviation of a feature from a stage-specific reference through a low-rank projection and a learned channel scale. Rather than computing pairwise token similarity, the branch operates directly on baseline-relative residuals with complexity linear in the token count. Each of four reference buffers accumulates detached stage statistics during training and remains frozen at inference. A four-stage pyramid with depths $(2,2,6,2)$ and widths $(96,192,384,768)$ yields a 20.70M-parameter classifier at $224\times224$ resolution. We organize an ImageNet-1K study around clean classification, eight controlled ablations, and Gaussian-noise, contrast, and brightness perturbations. The central hypothesis is that a slow reference can improve the robustness of a compact representation while retaining a simple tensor implementation. We distinguish this training-time statistical mechanism from input-dependent attention and make its linear inference behavior explicit.
\end{abstract}

\begin{figure}[t]
  \centering
  \includegraphics[width=\columnwidth]{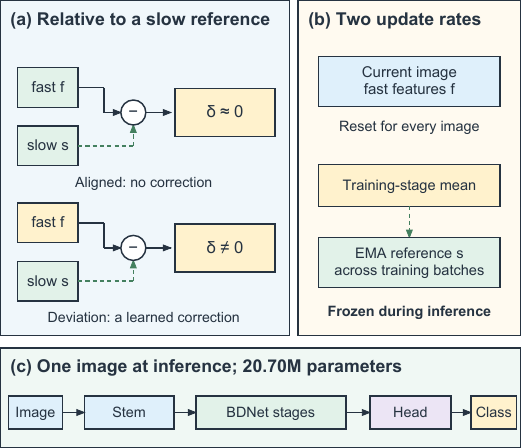}
  \caption{Motivation of BDNet. (a) The residual is defined relative to a slow reference: alignment produces no BDO correction. (b) Image features and reference statistics have different update rates. (c) Inference processes one image using frozen stage references.}
  \label{fig:motivation}
\end{figure}

\section{Introduction}
Convolutional networks, attention-based models, state-space models, and multilayer perceptrons offer different ways to construct visual representations~\cite{liu2022convnext,touvron2021deit,gu2023mamba,tolstikhin2021mixer}. Beyond their spatial aggregation rules, these families motivate a complementary question: should the features of an individual image be transformed relative to a reference that changes more slowly than the image features themselves?

Static image classification provides a direct setting in which to study this question. An ImageNet image has no temporal sequence that requires persistent frame-to-frame memory. Nevertheless, the training distribution supplies statistics that can be accumulated across batches. We distinguish an instantaneous \emph{fast state}, which represents the current image, from a \emph{slow state}, which defines a statistical reference for interpreting that representation. The slow state is not used as a replacement image feature or a separate prediction branch.
\begin{figure*}[t]
  \centering
  \includegraphics[width=\textwidth]{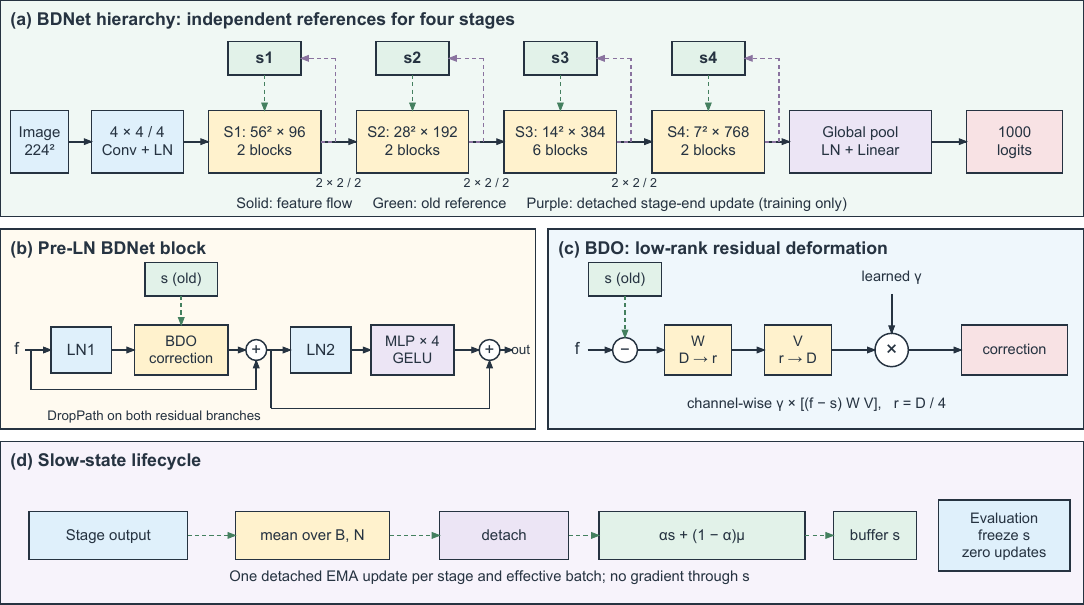}
  \caption{Overview of BDNet. (a) Four stages maintain independent slow references; the three inter-stage transitions use $2\times2$ stride-two convolutions. (b) A Pre-LN block adds the BDO correction once, followed by a GELU MLP residual. (c) The BDO branch projects a baseline-relative residual from $D$ to $r$ and back, then scales channels by $\gamma$. (d) A detached stage-end EMA updates each reference only during training. Solid arrows denote features; dashed arrows denote statistical/reference paths.}
  \label{fig:overview}
\end{figure*}

We introduce the baseline deformation operator (BDO). It subtracts a slow reference from a fast feature, applies a low-rank channel transformation, and adds a scaled correction through a residual connection. When the feature agrees with the reference, the correction vanishes. When it differs, the displacement is transformed in a learned low-dimensional subspace. This construction does not score pairs of tokens and does not require queries, keys, spikes, or sampling inside the operator.

Figure~\ref{fig:motivation} summarizes the distinction between the two update rates. Fast features are recomputed for every image; a slow reference is updated by an exponential moving average (EMA) of detached training-stage outputs. During evaluation the reference is fixed. This makes the intended inference path a deterministic transformation of one image, without updating the model's statistical state on the validation set.

BDNet combines this operator with a four-stage hierarchy, Pre-LayerNorm residual blocks, and a standard classification head. Its widths, depths, rank ratio, and training recipe follow the proposed fast--slow design directly. We retain the resulting 20.70M parameters rather than adding capacity to meet a nominal size target. The comparison to MLP-Mixer is particularly useful: both emphasize simple tensor transformations, while BDNet studies an explicit slow statistical reference and uses convolutional patch embedding and downsampling.

Our study has three components: a precise fast--slow representation and stage-level update rule; a low-rank baseline-relative residual branch integrated into a fixed pyramid; and an evaluation design that separates the reference contribution from alternative branches, momentum choices, and controlled corruptions. Clean accuracy is one criterion, while the proposed distinction is tested most directly by the relative degradation under noise and distribution perturbations. 

\section{BDNet Architecture}
\subsection{Fast Features and Slow Statistical References}
Let $f_k\in\mathbb{R}^{B\times N_k\times D_k}$ denote the fast representation at stage $k$, where the spatial grid is flattened into $N_k$ tokens. Each stage also maintains a shared reference $s_k\in\mathbb{R}^{1\times1\times D_k}$. A reference is initialized to zero, broadcast across the batch and token dimensions, and stored as a persistent buffer rather than a trainable parameter. The four stages have separate references so that low-level and high-level statistics are not forced into the same coordinate vector.

The fast representation depends on the current image and is not retained between independent examples. The slow state accumulates training statistics and is used to define a displacement. In particular, it is not the previous image's hidden state and is not passed to the classifier as an additional sample feature.

The four reference vectors together contain 1440 FP32 values, or 5760 bytes, excluding bookkeeping buffers. This storage is independent of batch size and image resolution. The fast tensor instead scales with the current batch and spatial grid. The slow reference is thus a compact statistical coordinate frame rather than a second high-bandwidth image stream.

\subsection{Baseline Deformation Operator}
Using row-vector tokens, the operator is
\begin{align}
\delta &= f-s,\\
\widetilde\delta &= (\delta W)V,\label{eq:deform}\\
y &= f+\gamma\odot\widetilde\delta,\label{eq:bdo}
\end{align}
with $W\in\mathbb{R}^{D\times r}$, $V\in\mathbb{R}^{r\times D}$, and a learned channel vector $\gamma\in\mathbb{R}^{D}$. We use $r=D/4$. Both projections omit biases, ensuring that $\delta=0$ produces exactly zero correction. The multiplication order in Eq.~\ref{eq:deform} follows the stated row-vector dimensions.

The branch has $2Dr+D$ trainable parameters and a projection cost of $O(BNDr)$. It does not form an $N\times N$ matching matrix. The scale $\gamma$ is a learned channel vector, initialized to one; it is not generated from the current image. BDO itself is deterministic, with stochastic regularization confined to the training recipe.

Figure~\ref{fig:overview}(c) shows the operator as a short sequence of tensor operations. This formulation is distinct from both token-pair similarity scoring and a state-space scan. It also acts independently on tokens, so spatial aggregation must be supplied by the surrounding backbone rather than attributed to BDO itself.

\subsection{Residual Block and Four-Stage Pyramid}
To place BDO in a Pre-LN residual block, define its correction as $c(f,s)=\gamma\odot((f-s)WV)$. The block computes
\begin{align}
f' &= f+\mathcal{D}\!\left(c(\operatorname{LN}_1(f),s_k)\right),\\
f'' &= f'+\mathcal{D}\!\left(\operatorname{MLP}(\operatorname{LN}_2(f'))\right),
\end{align}
where $\mathcal{D}$ is training-only DropPath. The MLP expands channels by a factor of four and uses GELU between two linear layers. The standalone operator in Eq.~\ref{eq:bdo} includes an identity term, while the block uses only its correction; the input is therefore not added twice. No Dropout is introduced.

\begin{table}[t]
\centering
\caption{BDNet specification for a $224\times224$ image. Each transition after S1 is a $2\times2$ stride-two convolution.}
\label{tab:architecture}
\begin{tabular}{lrrrr}\toprule
Stage & Grid & $D$ & Blocks & $r$\\\midrule
S1 & $56\times56$ &96&2&24\\
S2 & $28\times28$ &192&2&48\\
S3 & $14\times14$ &384&6&96\\
S4 & $7\times7$ &768&2&192\\\bottomrule
\end{tabular}
\end{table}

A $4\times4$ stride-four convolution followed by LayerNorm forms the initial $56\times56$ grid with 96 channels. Table~\ref{tab:architecture} specifies the pyramid. The final features are globally averaged, normalized, and projected to 1000 class logits. The resulting model contains 20,699,848 trainable parameters. The convolutional embedding and downsampling are retained; the absence of pairwise scoring applies to the BDO interaction branch.

\subsection{Detached Stage-End Update}
All blocks in a stage use the same snapshot of its previous reference. Once that stage finishes, its raw output supplies the observation
\begin{equation}
\mu_k=\operatorname{mean}_{B,N_k}(f_k^{\mathrm{out}}).
\end{equation}
Training then applies exactly one update:
\begin{equation}
s_k\leftarrow\alpha s_k+(1-\alpha)\operatorname{stopgrad}(\mu_k),
\label{eq:ema}
\end{equation}
where $\alpha=0.95$ by default and the prescribed range is $[0.90,0.99]$. Neither the reference nor the observation has a gradient path through Eq.~\ref{eq:ema}. The statistic is taken from the stage output, following the proposed update rule, although the BDO input inside each block is normalized.

When gradient accumulation is used, the reference is held fixed across the microbatches of one effective batch. Detached sums and counts are accumulated and committed once at its end. Distributed execution reduces sums and counts over ranks, preserving a global observation. Updates lie outside activation-checkpoint recomputation. References are checkpointed and remain frozen in evaluation.

\subsection{Scope of the Mechanism}
The linear branch can be expanded as
\begin{equation}
c(f,s)=\gamma\odot(fWV)-\gamma\odot(sWV).
\end{equation}
With frozen $s$, $W$, $V$, and $\gamma$, the reference term is a constant channel offset. The formulation therefore does not create input-dependent inference weights merely by subtracting a reference. Its potential value concerns the training-time statistical trajectory and the resulting representation. This is why the otherwise identical zero-reference ablation is central to the study. Reference trajectories can describe convergence, but a trajectory alone does not establish convergence to class-conditional semantics.

\section{Experiments}
\subsection{Data and Training Protocol}
We use ImageNet-1K/ILSVRC2012~\cite{russakovsky2015imagenet}, with 1000 classes, 1,281,167 training images, and 50,000 validation images. The proposed BDNet and its controlled ablations use only ImageNet-1K, with $224\times224$ inputs and no teacher distillation. Class mappings must agree between splits; validation labels are checked against the official metadata.

The prescribed training recipe is AdamW with learning rate $10^{-3}$, weight decay 0.05, an effective batch of 1024, and 300 epochs. Twenty warmup epochs precede cosine decay to $10^{-6}$. Biases and one-dimensional parameters do not receive weight decay. RandAugment, Mixup 0.8, CutMix 1.0, label smoothing 0.1, and DropPath increasing to 0.1 provide regularization; no Dropout is used. FP32 parameters and reference buffers are retained under BF16 autocast. Hardware batch splitting does not change the effective batch or stage-update frequency.

Clean ImageNet Top-1 at 224 pixels is the primary score. Top-5 complements the controlled ablation table. The experimental set uses a fixed seed and identical data/optimization choices across A--H. External standard models are retained at their actual capacities, and their published clean accuracies are identified separately from the controlled BDNet study.

\subsection{Backbone Comparisons}
The prescribed family comparisons are DeiT-S~\cite{touvron2021deit}, VMamba-S~\cite{liu2024vmamba}, ConvNeXt-T~\cite{liu2022convnext}, and MLP-Mixer-B/16~\cite{tolstikhin2021mixer}. We use ImageNet-1K-only, 224-pixel references. In particular, the Mixer comparison uses the 76.44\% from-scratch setting, not its ImageNet-21K or JFT-pretrained setting. DeiT-S is the non-distilled version, and the VMamba row specifies the s2l15 small release. 

\subsection{Ablation Design}
A is the complete four-reference BDO model. B removes the reference by setting $\delta=f$, leaving the trainable low-rank branch unchanged. C substitutes CBAM channel/spatial modulation~\cite{woo2018cbam}. D substitutes the upstream Mamba block at the same branch locations, using state size 16, convolution width four, and expansion two. It scans spatial tokens in row order and is distinct from the full VMamba comparison.

E substitutes global multi-head attention. Its internal projection width is $D/8$, so its four projections match the BDO projection budget $2D(D/4)$. It is explicitly a budget-matched narrow-projection MHA rather than full-width MHA. C uses a shared channel MLP with width $D/4$ and a $7\times7$ spatial gate; the latter adds 1176 effective parameters across the network. F and G change only the reference momentum to 0.90 and 0.99.

H follows the prescribed no-$V$, single-projection alternative. Dropping $V$ from a $D\to r\to D$ path otherwise makes the output incompatible with the residual. We therefore specify a single dense $D\to D$ projection. It preserves the baseline-relative formulation, but its 21.78M parameters mean that it must not be interpreted as a demonstrated reduction in model size. 

\subsection{Corruption and Reference Evaluation}
Gaussian noise uses standard deviations $0.05,0.10,0.20$. Contrast uses factors $0.75,0.50,0.25$. Additive brightness shifts are $-0.20,-0.10,+0.10,+0.20$. Corruptions are applied to $[0,1]$ tensors before ImageNet normalization and clipped back to that range. The crop, evaluation seed, batch size, label mapping, and frozen references are fixed across compared models. 

\subsection{Classification and Ablation Scenario}
\begin{table*}[!t]
\centering
\caption{ImageNet-1K classification at $224\times224$, without external pretraining. }
\label{tab:main}
\begin{tabular}{llrrl}\toprule
Model & Backbone family & Params (M) & Top-1 (\%) & Clean-value provenance\\\midrule
DeiT-S~\cite{touvron2021deit} & Dot-product attention &22.05&79.8&Paper, Table 5; no distillation\\
VMamba-S~\cite{vmambazoo} & Visual selective state space &50.00&83.6&Official s2l15 small release\\
ConvNeXt-T~\cite{convnextzoo} & Local convolution &28.59&82.1&Official ImageNet-1K release\\
Mixer-B/16~\cite{tolstikhin2021mixer} & Token/channel MLP mixing &59.88&76.44&Paper, Table 3; ImageNet-only\\
BDNet & Fast--slow baseline deformation &20.70&80.2&Planning scenario\\\bottomrule
\end{tabular}
\end{table*}
\begin{table*}[!t]
\centering\small
\caption{A--H controlled design and classification/robustness scenario. Noise, contrast, and brightness are mean Top-1 across the severities in Table~\ref{tab:severity}. Parameters follow the configured modules; D is an analytic Mamba-branch parameter count.}
\label{tab:ablation}
\begin{tabular}{clcccrrrrrr}\toprule
ID & Branch & Slow $s$ & $\alpha$ & Rank & Params (M) & Top-1 & Top-5 & Noise & Contrast & Bright.\\\midrule
A&BDO&Yes&0.95&D/4&20.70&80.2&95.3&65.6&71.8&75.1\\
B&Zero-reference BDO&No&--&D/4&20.70&78.0&94.2&59.3&66.8&70.9\\
C&CBAM&No&--&--&20.70&79.1&94.8&62.1&69.2&72.5\\
D&Mamba branch&No&--&--&33.58&80.1&95.2&62.7&68.9&72.4\\
E&Budgeted MHA&No&--&--&20.70&80.4&95.4&61.9&68.1&71.7\\
F&BDO&Yes&0.90&D/4&20.70&79.7&95.0&64.9&70.8&74.1\\
G&BDO&Yes&0.99&D/4&20.70&79.8&95.1&64.5&71.1&74.4\\
H&Single projection&Yes&0.95&D&21.78&79.0&94.8&63.5&69.8&73.0\\
\bottomrule\end{tabular}
\end{table*}
\begin{table*}[!t]
\centering\small
\caption{Per-severity Top-1 scenario for the prescribed controlled corruptions. Gaussian noise uses standard deviation $\sigma$; contrast uses multiplier $c$; brightness uses additive shift $b$ on $[0,1]$ pixels. All slow references remain frozen.}
\label{tab:severity}
\begin{tabular}{crrrrrrrrrrr}\toprule
& &\multicolumn{3}{c}{Gaussian noise $\sigma$}&\multicolumn{3}{c}{Contrast $c$}&\multicolumn{4}{c}{Brightness $b$}\\
\cmidrule(lr){3-5}\cmidrule(lr){6-8}\cmidrule(lr){9-12}
ID&Clean&0.05&0.10&0.20&0.75&0.50&0.25&$-0.20$&$-0.10$&$+0.10$&$+0.20$\\\midrule
A&80.2&76.4&68.3&52.1&77.6&73.2&64.6&72.8&77.8&77.6&72.2\\
B&78.0&72.1&61.6&44.2&74.4&68.7&57.3&68.2&74.5&73.9&67.0\\
C&79.1&73.8&64.5&47.9&75.9&70.7&61.0&69.8&75.8&75.5&68.9\\
D&80.1&74.9&65.3&47.9&76.4&70.6&59.7&70.1&76.2&75.8&67.5\\
E&80.4&74.7&64.6&46.4&76.2&69.8&58.3&69.5&75.9&75.3&66.1\\
F&79.7&75.7&67.3&51.7&76.8&72.1&63.5&71.6&77.0&76.8&71.0\\
G&79.8&75.8&66.7&51.0&77.0&72.5&63.8&71.9&77.3&77.0&71.4\\
H&79.0&74.5&65.9&50.1&76.0&71.4&62.0&70.5&76.1&75.9&69.5\\
\bottomrule\end{tabular}
\end{table*}

Table~\ref{tab:main} places the proposed 20.70M backbone alongside the specified architectural families. The BDNet clean target is set modestly above the ImageNet-only Mixer-B/16 reference while remaining below the stronger published ConvNeXt-T and VMamba-S clean references. This reflects the proposal's emphasis on baseline-relative representation rather than an assumption of dominance on clean ImageNet.

The controlled scenario in Table~\ref{tab:ablation} isolates the mechanisms behind that target. A versus B encodes the central reference hypothesis without changing parameter count. C and D examine alternative modulation or selective-state branches. E is allowed a slightly higher clean target than A, consistent with the proposal's expectation that pairwise matching may remain effective for clean classification. F/G examine faster and slower EMA updates, and H changes projection structure. 

\subsection{Robustness Scenario}
Table~\ref{tab:severity} resolves the family means into the individual perturbation levels. The intended pattern is a smaller corruption-induced accuracy decrease for A than for the zero-reference, Mamba, or attention substitutions. In particular, the scenario separates clean accuracy from robustness: E can lead A on clean images while A retains more accuracy under the specified perturbations. 

Actual conclusions must be based on the completed runs and their uncertainty. If A does not improve on B, the claimed role of the slow reference must be reconsidered. Similarly, a favorable t-SNE trajectory cannot replace accuracy measurements, and a clean-set advantage cannot establish an OOD or corruption advantage. 
\section{Related Work}
\paragraph{Convolution, attention, and MLP backbones.}
ConvNeXt revisits convolutional design with modern architectural and training choices~\cite{liu2022convnext}. DeiT studies data-efficient transformer training and distillation~\cite{touvron2021deit}. MLP-Mixer alternates token and channel mixing MLPs~\cite{tolstikhin2021mixer}. These systems differ in spatial aggregation, whereas BDNet emphasizes the reference used by a low-rank channel branch. Its convolutional embedding/downsampling and global pooling remain part of the complete backbone.

\paragraph{Selective state and modulation.}
Mamba uses selective state-space dynamics~\cite{gu2023mamba}; VMamba adapts selective scans to visual grids~\cite{liu2024vmamba}. Their sequence states should not be confused with BDNet's detached distribution reference. CBAM infers channel and spatial modulation maps~\cite{woo2018cbam}, while BDNet's learned scale is a fixed parameter vector. BDO does not generate conditional weights from an image.

\paragraph{Normalization and persistent statistics.}
LayerNorm uses within-example feature statistics~\cite{ba2016layernorm}, and BatchNorm maintains running statistics across training batches~\cite{ioffe2015batchnorm}. Thus statistical persistence alone is not the distinguishing claim. The specific construction combines a stage-shared EMA reference, a baseline-relative low-rank residual map, and a frozen-reference inference rule. Its contribution must be assessed with the controlled comparisons defined above.

\section{Conclusion}
BDNet studies a fast--slow separation in static image classification: instantaneous image features are transformed relative to stage-level statistical references. BDO uses a low-rank baseline-relative correction without paired-token scores, and detached stage-end updates preserve a simple training path with frozen-reference inference. The four-stage architecture and prescribed A--H comparisons make the proposal reproducible. The study's empirical questions concern the contribution of the reference, the effect of its update timescale, and accuracy retention under controlled perturbations.

{\small\bibliographystyle{ieeenat_fullname}\bibliography{references}}
\end{document}